\documentclass{article}
\usepackage{lipsum,spconf,multicol,times, epsfig, graphicx, amsmath, amssymb, floatrow, booktabs, url, multirow, comment,subfigure,mwe}
\usepackage{textcomp}

\usepackage[font=small,skip=1pt]{caption}

\graphicspath{ {./figs/} }
\usepackage[T1]{fontenc}
\def\etal{{\em et al. }}

\title{Making Third Person Techniques Recognize First-Person Actions in Egocentric Videos}
\name{Sagar Verma \quad Pravin Nagar \quad Divam Gupta \quad Chetan Arora}
\address{IIIT Delhi}

\begin{document}

\maketitle
\begin{abstract}
We focus on first-person action recognition from egocentric videos. Unlike third person domain, researchers have divided first-person actions into two categories: involving hand-object interactions and the ones without, and developed separate techniques for the two action categories. Further, it has been argued that traditional cues used for third person action recognition do not suffice, and egocentric specific features, such as head motion and handled objects have been used for such actions. Unlike the state-of-the-art approaches, we show that a regular two stream Convolutional Neural Network (CNN) with Long Short-Term Memory (LSTM) architecture, having separate streams for objects and motion, can generalize to all categories of first-person actions. The proposed approach unifies the feature learned by all action categories, making the proposed architecture much more practical. In an important observation, we note that the size of the objects visible in the egocentric videos is much smaller. We show that the performance of the proposed model improves after cropping and resizing frames to make the size of objects comparable to the size of ImageNet's objects. Our experiments on the standard datasets: GTEA, EGTEA Gaze+, HUJI, ADL, UTE, and Kitchen, proves that our model significantly outperforms various state-of-the-art techniques.
\end{abstract}


\section{Introduction}
\label{sec:intro}

With the improvement in technology and usability, wearable cameras like GoPro \cite{GoPro}, Pivothead \cite{pivothead}, and Microsoft Sensecam \cite{sense_cam} are becoming ubiquitous. These cameras are typically harnessed to a wearer's head giving the first-person perspective. We refer to such cameras as egocentric cameras. The unique perspective of the egocentric camera, as well as, the commonly available always-on feature, makes use of such cameras compelling in applications like extreme sports, law enforcement, lifelogging, home automation and assistive vision.

\begin{figure}[t]
\begin{center}
{\includegraphics[width=0.30\linewidth , height=20mm]{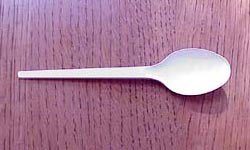}}
{\includegraphics[width=0.30\linewidth, height=20mm]{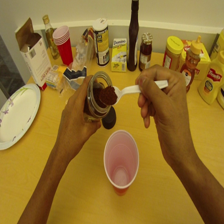}}
{\includegraphics[width=0.30\linewidth, height=20mm]{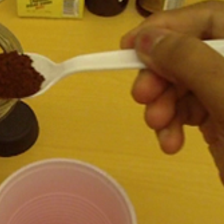}}
{\includegraphics[width=0.30\linewidth, height=20mm]{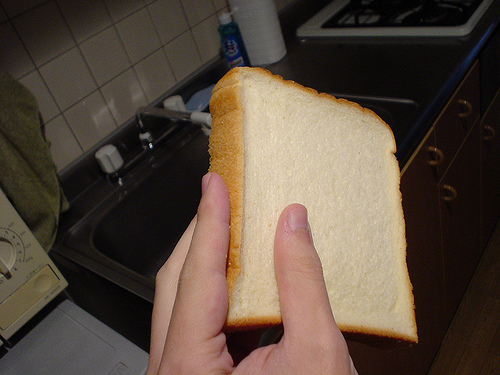}}
{\includegraphics[width=0.30\linewidth, height=20mm]{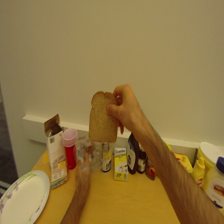}}
{\includegraphics[width=0.30\linewidth, height=20mm]{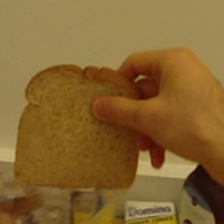}}
\end{center}
\caption{The top view nature of egocentric camera reduces the visible object size (second column) compared to that in ImageNet dataset \cite{russakovsky2015imagenet} (first column). This has lead to poor performance of RGB frame based action recognition in egocentric videos. We randomly crop $224\times224$ region from near the center and resize it to $300 \times 300$, which match the size of the object. This allows the proposed RGB stream and the overall model to achieve state-of-the-art performance for first-person action recognition.}
\end{figure}

The conventional third-person action recognition techniques use the pose of the actor as an important cue. However, the egocentric camera does not even see the actor or the wearer. Therefore, algorithms for first-person action recognition have typically relied on secondary cues such as wearer's hand motion, handled object attributes and camera ego-motion. Here, sharp changes in the viewpoint due to head motion, occlusion of the objects due to wearer's hands, and unconstrained environment have posed significant challenges for first-person action recognition.

Our focus in this work is to recognize wearer's action from an egocentric video. Unlike most of the state-of-the-art, our objective is to develop a generic feature learning technique for all types of action categories: actions involving hand-object interaction (e.g., \lq{take}\rq, \lq{pour}\rq, \lq{spread}\rq, \lq{stir}\rq, etc.), actions involving no hand-object interaction (e.g., \lq{walk}\rq, \lq{run}\rq, etc.), short-term action (e.g., \lq{fold}\rq, \lq{put}\rq, etc.) and long-term action (e.g., \lq{spread}\rq, \lq{stir}\rq, etc.).

The specific contributions of this paper are as follows:
\begin{enumerate}
\item We posit that deep neural network (DNN) models trained on third person videos do not adapt to egocentric actions due to the large difference in size of the objects visible in the two kinds of videos. We observe significant performance improvement in the standard models on increasing the object size and making them comparable to the size of objects typically found in the ImageNet dataset.
\item We propose several other minor contributions such as curriculum learning to handle first-person actions, such as `open' and `close', which are similar but opposite to each other.
\item Finally, in a significant departure from current state-of-the-art, we propose a single DNN model, that can dynamically adapt to all categories of egocentric actions. The proposed framework achieves state-of-the-art performance on various publicly available standard datasets for first-person action recognition, where category specific models, have been employed, confirming the truly generic nature of the proposed architecture.
\end{enumerate}

\section{Related Work}
\label{sec:format}

Conventional third-person action recognition techniques typically learn and match visual features from video frames based on key-points and descriptors \cite{ Willems2008, Lior2009}. Recently the methods using appearance and motion information around densely sampled point trajectories \cite{wang2013action, mihir_cvpr13, kraft2014accv}, as well as deep learned features have also been proposed with promising results \cite{karpathy2014action,   tran_c3d, twostream2014action}.

Most of the earlier works on first-person action recognition use hands and objects as important cues \cite{  fathi2011learning, pirsiavash2012detecting, ryoo2013first}. While earlier works focused on global features and IMU data \cite{spriggs2009temporal}, \cite{pirsiavash2012detecting} proposed an object-centric approach. McCandless and Graumann [12] introduced spatiotemporal pyramid histograms of objects appearing in the action. Recently, \cite{kitani2011fast} have suggested motion based histograms, \cite{suriya2016trajfeatures}, trajectory aligned features. Flow-based features have also been used by \cite{poleg2014temporal} and \cite{poleg2016compact}, who have proposed cumulative displacement curves and compact CNN architecture respectively for recognizing first-person actions. Each of these works focuses on one specific category of actions only.

The work closest to us is \cite{videoclass2015ng}, where the authors have similarly used a CNN-LSTM model for third person actions. They have trained and tested their model on a huge dataset of 1 million sports video from 1000 action categories downloaded from YouTube. In our case, the datasets are much smaller making it imperative to curate the features. We resize the region of interest to match the size of objects in the egocentric dataset and third person benchmark datasets as explained in the last section.

\section{Proposed Approach}

\begin{figure}[t]
\includegraphics[width=1.0\linewidth]{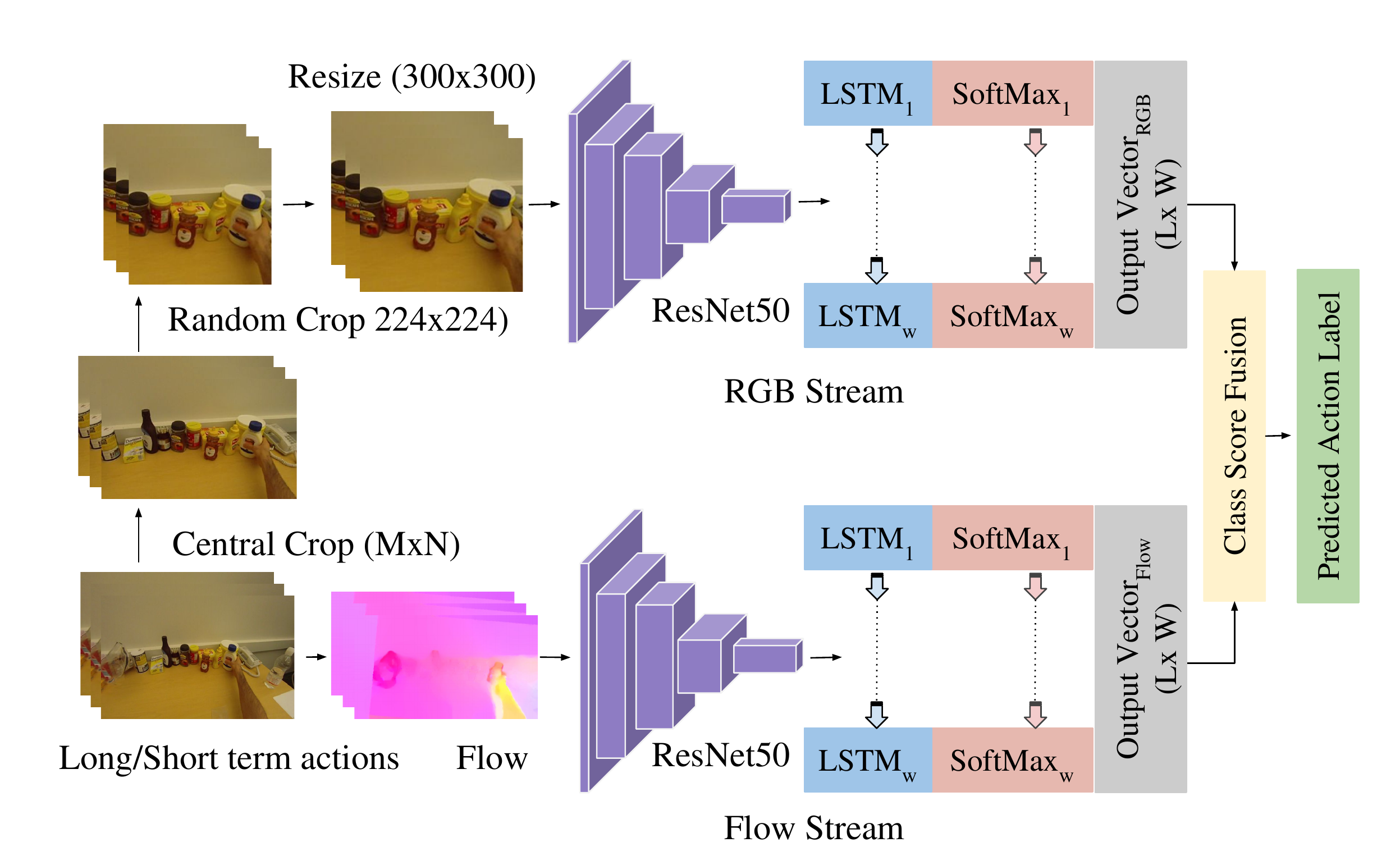}
\caption{Proposed Architecture}
\label{fig:res}
\end{figure}

Our emphasis is to learn generic features from egocentric videos. Learning visual representation, as well as, the motion patterns from scratch in a single network can make it excessively large requiring a huge training dataset. Given the scarcity of publicly available egocentric videos, we have adapted the transfer learning and data augmentation approach for our problem. The two stream architecture uses two modalities, namely RGB and optical flow for feature extraction. These extracted features are given as an input to LSTM. The details of proposed network model and the inputs are given below.

\paragraph*{Network Input: RGB frames and optical flow:}

Keeping in-line with our objective of the end to end training, we use RGB frames as an input to a pre-trained CNN. Only random cropping is used for data augmentation. We cropped $224 \times 224$ region from $M \times N$ central cropped raw image to increase the training set. The central crop dimension varies according to the datasets. To increase the object size for RGB stream, we have resized the cropped input from $224 \times 224$ to $300 \times 300$ so that objects size matches to the size of ImageNet's objects. To incorporate sequential information in the prediction process, we select a splice of size $W$ around each frame.

Motion patterns, as indicated by optical flow in an image are important cues for first-person actions in the proposed architecture. However, unlike some of the earlier works \cite{poleg2014temporal, poleg2016compact} using sparse optical flow, we propose to use dense optical flow \cite{liu2009beyond}, as some of the fine object manipulation activities are hard to capture in a sparse scenario. Wearer's head is often the dominant source of flow in an egocentric video, but is often unrelated to the action being performed. Therefore, as suggested in \cite{suriya2016trajfeatures, suriya2016cvpr}, we pre-process the optical flow to compensate the component due to head movement. We achieve this by canceling frame to frame homography.    

\paragraph*{Architecture:}

We use two streams in the proposed model, one using RGB frame as the input and other using optical flow. Both the streams use the same architecture (but trained on different inputs: RGB, Optical Flow).

We have experimented with two well known convolutional neural network(CNN) models, namely VGG-16 \cite{vgg2014} and ResNet-50 \cite{resnet50} pre-trained on the ImageNet dataset. We fine-tune the models on the egocentric data and extract 2048 dimensional feature vector from the second fully connected layer to be given as an input to the LSTM module. We have selected ResNet-50 over VGG-16 because of its better empirical performance on our dataset.

\renewcommand{\tabcolsep}{0.05cm}
\begin{table}[t]
\centering
\begin{tabular}{l c c c c c c}
\toprule[0.2mm]
\textbf{Dataset} & \textbf{Subjects} & \textbf{Frames} & \textbf{Classes} &  \multicolumn{2}{c}{\textbf{Accuracy}} \\
& & & & \textbf{Current} & \textbf{Ours}\\
\midrule
GTEA \cite{fathi2011learning} & 4 & 31,253 & 11 & 68.50\cite{suriya2016cvpr} & 82.71 \\
EGTEA+ \cite{fathi2011learning} & 32 & 1,055,937 & 19 & NA & 66 \\
Kitchen \cite{spriggs2009temporal} & 7 & 48,117 & 29 & 66.23\cite{suriya2016cvpr} & 71.92 \\
ADL \cite{pirsiavash2012detecting} & 5 & 93,293 & 21 & 37.58\cite{suriya2016cvpr} & 44.13 \\
UTE \cite{lee2012discovering} & 2 & 208,230 & 21 & 60.17\cite{suriya2016cvpr} & 65.12 \\
HUJI \cite{poleg2016compact} & NA & 1,338,606 & 14 & 86\cite{poleg2016compact} & 93.92 \\
\bottomrule[0.2mm]
\end{tabular}
\caption{Accuracy comparison of our method with state-of-the-art and statistics of egocentric video datasets}
\label{table:datasets}
\end{table}

To incorporate temporal information from the video, we use one layer of LSTM units, which takes $2048$ dimensional input from the CNN. The LSTM is unrolled $W$ times to include long-term temporal dependency. The weights for all the CNNs supplying input to the LSTMs have been tied to keep the control on overall trainable parameters.

The output of the LSTM module from each stream depends on the number of input frames given to it. If $W$ frames are given as an input, the output of each stream is $L \times W$ dimensional vector, where $L$ represents the number of labels or action classes. It has been shown in the earlier works that combining outputs of unrolled LSTM units in various ways (first/ last/ average/ max-frequency) does not affect the accuracy in any significant way. In the proposed model, we take the weighted average of $L$ dimensional vectors produced by each unrolled unit to report the final result.

\section{Experiments and results}

\begin{figure}[t]
    \begin{center}
        {\includegraphics[width=0.3\linewidth, height=0.25\linewidth]{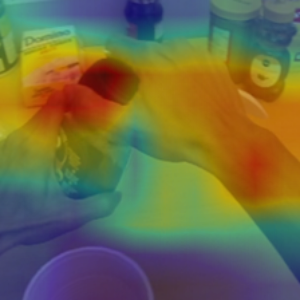}} \;
        {\includegraphics[width=0.3\linewidth, height=0.25\linewidth]{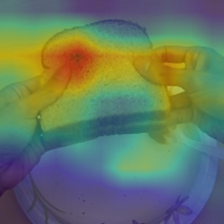}} \;
        {\includegraphics[width=0.3\linewidth, height=0.25\linewidth]{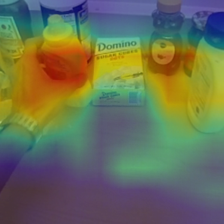}} \\ \vspace{0.5em}
        {\includegraphics[width=0.3\linewidth, height=0.25\linewidth]{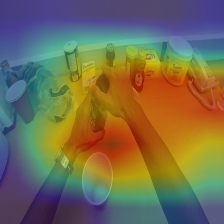}} \;
        {\includegraphics[width=0.3\linewidth, height=0.25\linewidth]{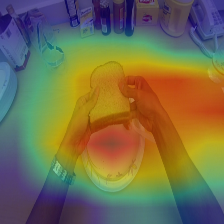}} \;
        {\includegraphics[width=0.3\linewidth, height=0.25\linewidth]{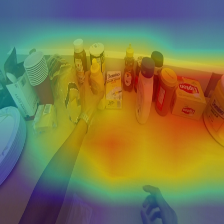}}
    \end{center}
    \caption{We use Gradient-weighted Class Activation Mapping (Grad-CAM) for visualization of important regions \cite{selvaraju2016grad}. Top and bottom rows show visualization of resized and normal inputs respectively for the open, put and take action classes (columnwise). From visualization it's evident that after increasing the object size and make them comparable to size of ImageNet's objects the emphasis on the objects inscreases, that in turn leads to improvement in accuracy of RGB stream.  }
\end{figure}

\paragraph*{Datasets:}

For our experiments on actions involving hand-object interaction we have used four different publicly available datasets: GTEA \cite{fathi2011learning}, Kitchen \cite{spriggs2009temporal}, ADL \cite{pirsiavash2012detecting} and UTE \cite{lee2012discovering}. For ADL \cite{pirsiavash2012detecting} and UTE \cite{lee2012discovering} datasets, we use the annotations provided by Singh \etal \cite{suriya2016cvpr}, who have annotated a subset of the original dataset. Other parts of the video are simply labeled as 'background'. For testing on actions involving no hand-object interaction, we use HUJI dataset \cite{poleg2016compact}. It consists of 14 action classes. The dataset is evenly distributed among different classes. Each class consists of many hours of videos containing a total of 82 hours of annotated data. Table \ref{table:datasets} summarizes statistics of various datasets used in our experiments. It is important to note that unlike state-of-the-art, we use a single architecture, which is used for all the datasets and action categories.

\paragraph*{Evaluation methodology:}

We use leave-one-subject-out policy for training and validation and report classification accuracy on the unseen test subjects. As described in the previous section we use a splice of $11$ frames input. For frame-wise prediction, we use the predicted class of the splice as the prediction of all the frames in the splice. All the reported accuracy numbers have been computed frame wise, consistent with the state-of-the-art.

\renewcommand{\tabcolsep}{0.05cm}

\begin{table}
\centering
\begin{tabular}{l c c c c} \toprule[0.2mm]
\multirow{2}{*}{\textbf{Stream}} & \multicolumn{4}{c}{\textbf{Frame level Accuracy}} \\ \cmidrule{2-5}
& \textbf{GTEA} \cite{fathi2011learning} & \textbf{Kitchen} \cite{spriggs2009temporal}  & \textbf{ADL} \cite{pirsiavash2012detecting} & \textbf{UTE} \cite{lee2012discovering} \\ \midrule
RGB & 81.93 & 62.23 & 43.94  & 59.10 \\
Flow & 82.67 & 69.90 & 38.43 & 64.78 \\ \midrule
Combined & 82.71  & 71.92 & 44.13 & 65.12 \\ \bottomrule[0.2mm]
\end{tabular}
\caption{Analysis of the proposed model using only RGB, flow and combined input.}
\label{table:flowVsrgb}
\end{table}

\paragraph*{Implementation details:}

To fine-tune VGG-16 and ResNet-50 models, input frames are normalized by mean and variance computed over complete datasets. Learning rate of 0.001, momentum of 0.9, learning rate decay of 0.1, step-size of 10K iteration and weight decay of 0.005 are used. The model is trained for 50K iterations with a batch size of 128 images. For optical flow, we use the same batch size, weight decay, learning rate decay, momentum. We choose a step-size of 20K and perform training for 70K iterations.

For RGB we take the output of the second fully connected layer and give it to an LSTM unit with 1024 cells. There are 11 unrolled LSTM units, corresponding to 11 input frames, connected in a unidirectional manner. Base learning rate of 0.001, momentum of 0.9, learning rate decay of 0.1, step size of 50K iterations and weight decay of 0.005 are used for training the LSTM. For training on flow frames, we keep the momentum, learning rate decay, weight decay and base learning rate same as those of RGB model. Learning rate is decreased by one-tenth after every 20K iterations and training is stopped at 70K iterations.

\paragraph*{Results and discussion:}

\begin{figure}[t]
\includegraphics[width=0.95\linewidth]{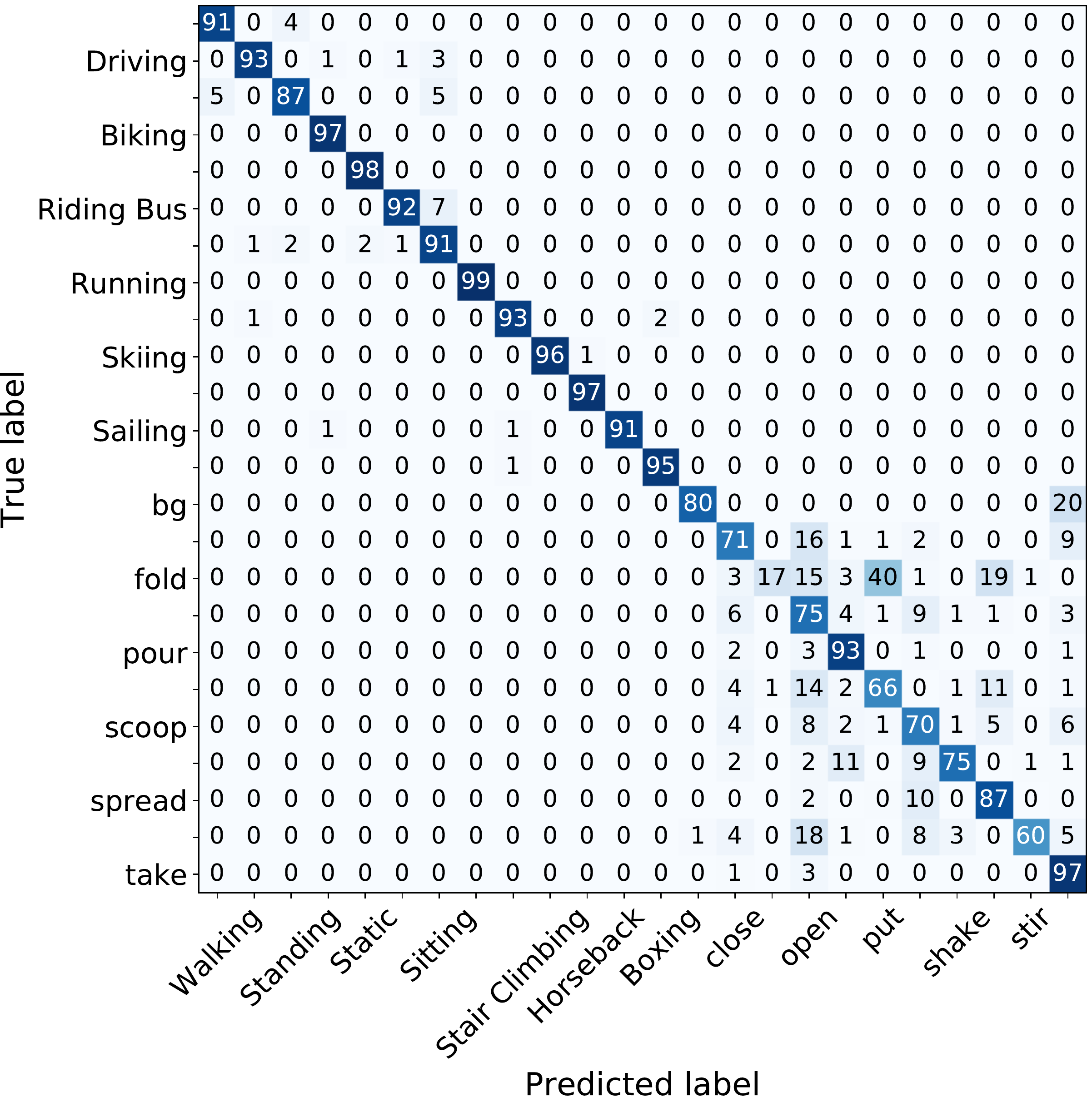}
\caption{Our model can be used for all kinds of actions together. Since most of the datasets contain only one kind, to validate the hypothesis, we trained our model after combining GTEA and HUJI datasets and got an accuracy of over $86.85\%$. Above, confusion matrix for the experiment shows that there is very little confusion between the action classes involving hand-object interaction (rows 14-24) and the ones without (rows 1-13). Confusion between `open' and `close' or `spread' and `put', are caused by action similarity and is present otherwise also in the state-of-the-art as well.}
\label{fig:mixed-actions}
\end{figure}
\renewcommand{\tabcolsep}{0.05cm}

We test our architecture with various temporal window sizes (5, 11, 15, and 21 frames) and chose 11 for its best performance empirically. Similarly, we have experimented with the different number of LSTM cells in our model: 128, 256, 512, 1024 and 2048 and found 1024 cells to be performing the best.

Table \ref{table:flowVsrgb} shows performance comparison using only RGB stream, only flow stream and after combining them over various datasets. We use \cite{liu2009beyond} to compute optical flow and convert it to RGB flow images \cite{baker2011database} after flow compensation. 

The top view nature of egocentric camera reduces the size of objects compared to the size of objects in the ImageNet dataset. Unlike the state-of-the-art, if we increase the object size, the performance of RGB stream in the proposed architecture becomes comparable to the flow stream. We also use curriculum learning approach during training, where we initially merge the opposite actions with similar visual and temporal information, such as open and close. This has also helped in making RGB stream in the proposed model achieve state-of-the-art accuracy alone. Table \ref{table:datasets} compares the performance with state-of-the-art. Both the streams, individually as well as jointly, improve state-of-the-art by a significant margin across all categories and datasets.

We note that \cite{videoclass2015ng} reports an accuracy of $65\%$ using single frame and $67\%$ using LSTM. We observe an accuracy of $80.66\%$ using single frame optical flow and $82.67\%$ using LSTM. For RGB frame, we achieve an accuracy of $80.97\%$ using single frame and $81.92\%$ using video sequence.

We also tested on long-term actions datasets. Here, we follow the evaluation strategy of \cite{poleg2016compact} and get an average recall rate of  $93\%$ against the $86\%$ reported by them. Their network confuses with actions where the flow is often ambiguous like \lq sitting\rq and \lq standing\rq. We believe that our CNN-LSTM network is able to perform better due to RGB frames as a feature. 

The current state-of-the-art has developed different techniques for long or short-term actions and with or without handled objects. However, in a real-life setting, a wearer is likely to be involved in a mixed action setting requiring a single model to be capable of recognizing all categories of action classes. To validate the applicability of our model for such a scenario, we have mixed the samples from GTEA and HUJI datasets. Figure \ref{fig:mixed-actions} gives the confusion matrix for the experiments. It is evident that the proposed network does not seem to have any confusion in the different category of actions and the confusing pairs, like shake-stir and fold-put are the usual ones where even current state-of-the-art has problems in disambiguating. The experiment indicates the much more practical applicability of the proposed technique.

\section{Conclusion}

Earlier works for first-person action recognition have explored various egocentric cues such as the motion of wearers head or hands and objects present in the scene and able to detect only a subset of all action categories. In this paper, we have proposed a CNN-LSTM model which can recognize all categories of action classes. A generic architecture, analysis in terms of visible object size, and curriculum learning exploiting similar action classes are some of the contributions of the proposed work. In future, we would like to use the capability to recognize long-term activities involving a sequence of hundreds of shorter actions.

\paragraph*{Acknowledgement:} Chetan Arora has been supported by Infosys Center for Artificial Intelligence and Visvesaraya Young Faculty Research Fellowship from Government of India. Pravin Nagar has been supported by Visvesaraya Ph.D. Fellowship from Government of India.

\bibliographystyle{IEEEbib}
\bibliography{ms}

\end{document}